\documentclass[conference]{IEEEtran}
\IEEEoverridecommandlockouts

\usepackage{graphicx}
\usepackage{subcaption}
\usepackage{amsmath,amssymb,amsfonts}
\usepackage{mathtools}
\DeclarePairedDelimiter\floor{\lfloor}{\rfloor}
\usepackage{tabu}
\usepackage{multirow}
\usepackage{hyperref}
\hypersetup{colorlinks,allcolors=black}
\usepackage{balance} 

\begin{document}

\title{

BanglaNet: Bangla Handwritten Character Recognition using Ensembling of Convolutional Neural Network

}

\author{\IEEEauthorblockN{Chandrika Saha}
\IEEEauthorblockA{\textit{Department of Computer Science} \\
\textit{Western University, London, Canada }\\
csaha@uwo.ca}
\and
\IEEEauthorblockN{Md Mostafijur Rahman}
\IEEEauthorblockA{\textit{Department of Electrical and Computer Engineering} \\
\textit{The University of Texas at Austin}\\
mostafijur.rahman@utexas.edu}

}

\maketitle

\begin{abstract}
Handwritten character recognition is a crucial task because of its abundant applications. The recognition task of Bangla handwritten characters is especially challenging because of the cursive nature of Bangla characters and the presence of compound characters with more than one way of writing. In this paper, a classification model based on the ensembling of several Convolutional Neural Networks (CNN), namely, BanglaNet is proposed to classify Bangla basic characters, compound characters, numerals, and modifiers. Three different models based on the idea of state-of-the-art CNN models like Inception, ResNet, and DenseNet are trained with both augmented and non-augmented inputs. Finally, all these models are averaged or ensembled to get the finishing model. Rigorous experimentation on three benchmark Bangla handwritten characters datasets, namely, CMATERdb, BanglaLekha-Isolated, and Ekush has exhibited significant recognition accuracies compared to some recent CNN-based researches. The top-1 recognition accuracies obtained are 98.40\%, 97.65\%, 97.32\%, and the top-3 accuracies are 99.79\%, 99.74\%, and 99.56\% for CMATERdb, BanglaLekha-Isolated and Ekush datasets respectively. 

\end{abstract}


\section{Introduction}
Being the fifth most popular native language, Bangla is a vastly spoken language worldwide \cite{basu2012handwritten}. It is the native language of Bangladeshi people and one of the most spoken languages of India. The dignified history of the Bangla language makes it even more significant. UNESCO declared 21st February as International Mother Language Day to remember the sacrifice the language martyrs of Bangladesh made in 1952.  Alphabets of the Bangla language consist of 50 basic characters (including 39 consonants and 11 vowels), 10 numerals, more than 10 modifiers, and more than 300 compound characters. Compound characters are two or more consonants combined. Classifying isolated offline Bangla characters is an indispensable task for building a complete Bangla Optical Character Recognition (OCR) system. 

\begin{figure}
\includegraphics[scale=.55]{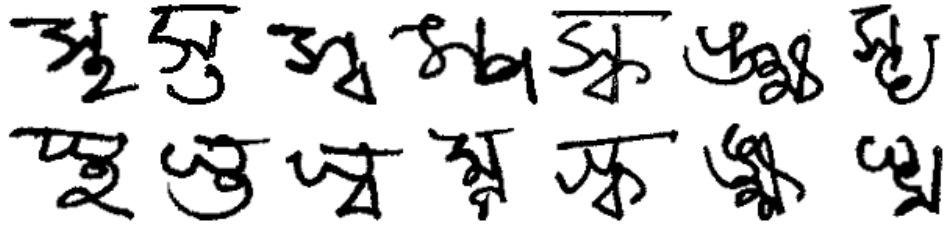}
\caption{Compound characters with multiple representations (same column represents same characters)} \label{fig1}
\end{figure}

Particularly, isolated offline handwritten character recognition is a challenging classification problem because a distinct human has a distinct style of writing. As a result, there are limitless possibilities of writing the same character. This task becomes even more intricate with the presence of complex shaped compound characters many of which have more than one way of writing. Fig.\ref{fig1} depicts some of such compound characters with more than one form. Moreover, some of the basic characters have striking resemblance; they differ by only a ‘matra’ (horizontal line above the character) or a dot below the character. Sometimes it's hard for even human agents to decode characters written by someone else. Therefore, to build a good enough Bangla Handwritten Character Recognizer (BHCR) a model that can properly identify isolated handwritten characters is a requisite.

For this type of pattern recognition task deep learning has become a popular choice. Specifically, the Convolutional Neural Network (CNN) discovered by Fukushima et al. \cite{fukushima1982neocognitron} is used for image-based classification tasks. It is described as a learning model that can classify images with geometrical similarities without the help of an additional feature extraction process. At the time of the publication, this type of classifier was not that popular because of the computational power required to train such models. But soon it started gaining attention and was implemented and tested by different researchers. LeCun et al. applied a simple CNN model to document-based recognition tasks \cite{lecun1998gradient}. A deeper and more complex CNN was proposed by Krizhevsky et al. \cite{krizhevsky2012imagenet} for the classification of the Imagenet dataset with 1000 classes \cite{deng2009imagenet}. Soon more complex CNN networks like ResNet \cite{he2016deep} and inception \cite{szegedy2015going} were implemented that gave state-of-art recognition accuracy for the very complex Imagenet dataset. 

The only problem with this type of network is, due to their very deep and complex structures a huge amount of computational power and time is needed to train these types of models which is necessary for very large datasets like Imagenet. For a shorter dataset, the size and complexity of a CNN can be minimized and a state-of-art accuracy can still be acquired. In this paper, three diverse CNN models inspired by several acclaimed models are applied to classify Bangla handwritten isolated characters on three distinguished Bangla handwritten character datasets.  At last, those three models are ensembled together to attain an aggravated accuracy and reduced loss. Datasets used for research are CMATERdb \cite{das2012statistical,das2012novel,das2009handwritten,das2015improved,dasbenchmark}, Banglalekha Isolated \cite{biswas2017banglalekha}, and Ekush \cite{rabby2018ekush}. These are some very recent and large datasets that are suitable to be applied to any deep learning framework. The recognition accuracies obtained for these datasets are, to the best of the author’s knowledge, the best accuracies so far.
\section{Literature Review}
\label{sec:lit}
Several researchers applied Convolutional Neural Networks for the classification of Bangla characters and digits. Nevertheless, most of the research work has been done on basic, compound consonants, and digits separately. Bangla modifiers are an integral part of the language which is used in almost every word, but surprisingly there has been a very sparse number of researches on recognizing them. However, some of the recent studies in the field of Bangla handwritten character recognition are portrayed in the following subsection. 

Rabby et al. proposed a CNN model ‘BornoNet’ for recognizing basic characters \cite{rabby2018bornonet}. They performed their experiment on the basic characters subset of datasets CMATERdb, BanglaLekha-isolated, and ISI \cite{bhattacharya2008handwritten}. Data augmentation was used to increase the number of training images and an automatic learning rate scheduler was used to tune the learning rate. As a regularization technique batch normalization and dropout are used in this research work. All these settings bring about a reasonable recognition rate.

Later on, after extending their study, Rabby et al. proposed a CNN-based model ‘EkushNet’ where they used a dataset of their construction \cite{rabby2018ekush}. The remarkable dataset ‘Ekush’ consists of more than three hundred thousand images of Bangla handwritten basic characters, modifiers, 52 predominant compound characters, and digits. Most of the experimental setup is almost identical to their previous research settings \cite{rabby2018bornonet}.

Alom et al. conducted an experiment on Bangla digit recognition based on Deep Belief Network (DBN) and different variations of CNN \cite{alom2017handwritten}. Researchers of this study applied three variations of the CNN model, CNN with dropout, with dropout and Gaussian filter, with Gabor filter and dropout, and concluded the last technique was the most effective one. For experimenting, CMATERdb 3.3.1 Bangla handwritten digit dataset was used. 

Alom et al. further the study by applying different state-of-art CNN models on handwritten digits, basic characters, and modifiers recognition separately \cite{alom2018handwritten}. CNN models like VGG Net, All Conv, Network in the network, ResNet, fractalNet, and DenseNet were applied to the CMATERdb dataset. This study concluded that DenseNet architecture accounted for the highest recognition correctness for handwritten digits, basic characters, and modifiers individually.

In their research Abir et al. constructed a CNN model with inception module followed by convolution and pooling blocks and ending with fully connected layers \cite{abir2019bangla}. They applied this diverse CNN model to BanglaLekha-Isolated dataset with 84 classes obtaining good precision.

In another research \cite{alif2017isolated} a tampered ResNet architecture was proposed which was trained and tested on BanglaLekha Isolated and CMATERdb dataset. Dropout and batch normalization are used in a single residual block in this proposed architecture. The model is tested for various configurations like different input image sizes, optimizers, etc. Remarkable accuracy was obtained for those two datasets.

In \cite{ashiquzzaman2017efficient} a DCNN model built with dropout and exponential linear unit activation was proposed for compound character recognition. On the CMATERdb 3.1.3.3 dataset, the proposed model obtained a good recognition rate.

In \cite{hakim2019handwritten} a DCNN model was proposed and trained only on Bangla basic characters and numerals of BanglaLekha Isolated dataset. For cross-validating the model, a new dataset was constructed. Although, the model is experimented with many different settings, for instance, varying learning rates, input size, number of epochs, batch sizes, and train validation split; the final result is somewhat unexplained.

Popular ResNet-50 model with transfer learning was used along with a modified one-cycle policy learning rate scheduler in \cite{chatterjee2019bengali}. Adam was used as an optimization technique \cite{kingma2014adam} and the model was trained with the BanglaLekha Isolated dataset with 84 classes of characters and digits. Remarkable accuracy was obtained with this transfer learning approach.

Chowdhury et al. evaluated the effect of data augmentation using only the basic characters from the BanglaLekha Isolated dataset \cite{chowdhury2019bangla}. The proposed CNN contains two fully connected layers followed by a 2 Conv-Pool block and SGD was used as an optimization technique. However, this study concluded that training with an augmented dataset provides much better accuracy than the original dataset.

In \cite{rizvi2019comparative} a comparative study between state-of-art machine learning technique and ResNet-18, a CNN model was performed on Bangla basic characters, digits, and compound characters. As datasets BanglaLekha Isolated and Ekush were combined and 15 classes of compound characters were selected. The results show that the CNN model outperformed the Support Vector Machine with a hybrid feature extraction-based technique. 

Noor et al. proposed an ensemble of two CNN models for recognizing noisy Bangla handwritten numerals \cite{noor2018handwritten}. Despite having data with tilted images, random box noises, and blurry images recognition rate was remarkable. 

In \cite{roy2017handwritten} one of the pioneering researchers of Bangla handwritten character recognition, Roy et al. proposed a greedy layer-wise CNN approach for Bangla compound character recognition. With the trained model a good recognition accuracy was obtained for CMATERdb 3.1.3.3.

Keserwani et al. proposed a two-phased DCNN scheme for Bangla compound character recognition \cite{hasan2019bangla}. Instead of randomly initializing weights of DCNN, they first trained a CNN to minimize the reconstruction loss and used weights gained from this phase to initialize weights of phase two. Adadelta optimization technique and ReLU activation were used while training the model.

From the aforementioned discussion, it is evident that Bangla isolated character recognition is explored mainly on basic characters, digits, and compound characters separately. Again, recognition of Bangla modifiers is rarely pursued. We need to include all these factors in a Bangla-isolated character recognizer as we will need to acquire a classifier that can identify most of the characters from a text including compound characters and modifiers. This paper proposed an ensembled CNN-based technique where three completely different types of CNN architectures are formed, trained on three prominent datasets, and at last, are ensembled to set up the final model.
\section{Methodology}
The formation of the proposed model, BanglaNet requires several steps that consist of data augmentation to increase the amount of data, batch normalization to normalize intermediate data, dropout to regularize the model, a proper learning rate scheduler, and the principles of inception, residual block, and dense net architecture. The following subsections provide a detailed overview of the techniques used to build the final model.  
\subsection{Data augmentation}
Because deep learning was developed to handle vast amounts of data that traditional machine learning algorithms are unable to handle, more data is essential to the field. Data augmentation is a process that is used to increase the data by altering a copy of the original data. Some augmentation techniques are mirroring, random cropping, zoom, rotation, shearing, local cropping, color shifting, Principle Component Analysis color augmentation, etc. Data augmentation increases the amount of data and brings variations in data which results in a robustly trained model as well as brings about a regularization effect. For handwritten character recognition, all the conventional augmentation strategies are not applicable as they might distort the image beyond recognition. However, in all the implementations of this paper rotation with a range of 10 degrees, shear ranging 10\%, zoom ranging 10\%, and width and height shift ranging 10\% is used. The same models are trained without any augmentation too; to inspect the effect of augmentation. 
\begin{figure*}[h!]
\centering
\includegraphics[scale=.95]{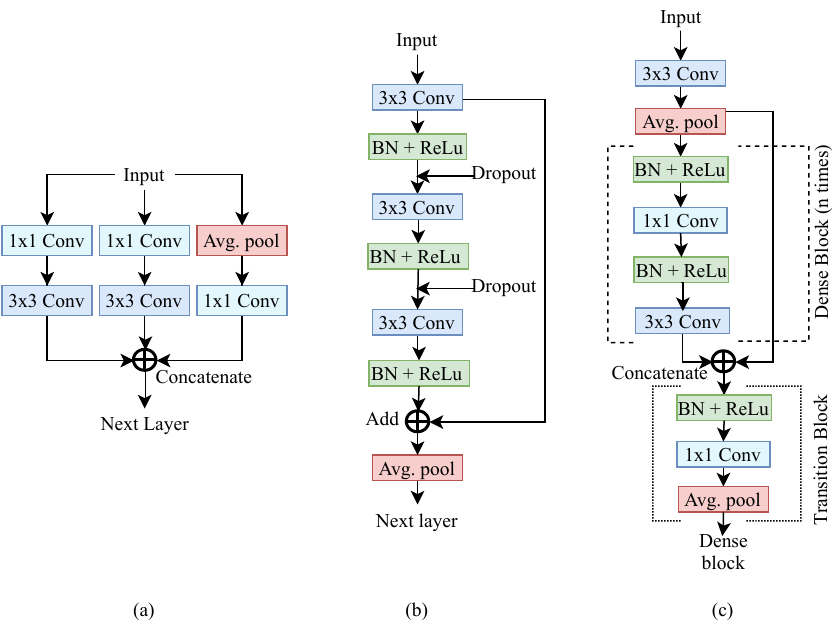}
\caption{(a) Inception block (b) Residual block (c)DenseNet block} \label{fig2}
\end{figure*}

\subsection{Batch normalization} 
Batch normalization is a technique used to maintain stable intermediate layers and provide CNN with a slight regularization effect \cite{ioffe2015batch}. Before starting the training, the input images are rescaled to have the input data within a range; this helps the model to perform better by making the model easily converge. Batch normalization is a similar technique used for hidden layers or intermediate layers of the deep learning model. The mathematical process of batch normalization is given in Equation (\ref{eq1}).
\begin{equation}
\label{eq1}
\scriptsize
\begin{split}
    \mu = \frac{1}{m}\sum_{i}^{}z\\
    \sigma^2 = \frac{1}{m}\sum_{i}{}{(z - \mu)}^2\\
    z_{norm} = \frac{(z - \mu)}{\sqrt{(\sigma^2 + \epsilon)}}\\
    \Tilde{z} = \gamma z_{norm} + \beta
\end{split}
\end{equation}

Here, $m$ is the size of the mini-batch, $z$ is the intermediate value of a hidden layer before applying the activation function, $\mu$ is mean, $\sigma^2$ denotes variance, $\epsilon$ is added to avoid a situation where the variance might be 0. This intermediate value $z$ is replaced with $\Tilde{z}$ during training time. The value of $z$ is first normalized by subtracting with mean and diving with the standard deviation. This normed value cannot be used directly as it will prevent the model from taking advantage of the non-linearity property of the activation function. So, two learnable parameters gamma and beta are added to modify the norm accordingly. The result is the intermediate values are normalized to have a specific mean and variance. Along with decreasing computational time, Batch normalization enables the learning model to be more robust towards covariate shift i. e., if the training and test images are from different distributions, the model will still be able to perform well as its’ parameters are normalized. For this paper batch normalization is used in all of the three models of consideration. 
\subsection{Dropout}
Dropout is an exoteric approach for regularization especially in deep learning \cite{JMLR:v15:srivastava14a}. This seemingly intelligible technique brings about the needed amount of regularization without increasing the computational cost immensely. Dropout works by setting a keep rate on nodes of any given layer. A node having a keep rate of 90\% has a 10\% possibility of being eliminated during the forward pass and the weights are not updated during the backward pass. This way the learning parameters are probabilistically obligated to take on more responsibility to make an accurate prediction. Thus, the overall testing loss is decreased. This technique is mainly used in the tailing of fully connected layers of a CNN architecture. But, in this paper, dropout is used in intermediate layers of a CNN model as well.
\subsection{Scheduled learning rate}
Learning rate is the most important hyperparameter for training any classifier. An excessive learning rate can make the model not converge at all while too low a learning rate can make training slow. To ensure a finer converges, an adaptive learning rate, namely, step decay is used for training the model. In step decay, the learning rate is halved after each five epochs thus ensuring an accurate global minima. The mathematical formula for step decay is given in Equation (\ref{eq2}).
\begin{equation}\label{eq2}
\scriptsize
lr = lr_0 * drop\_rate^{\floor*{\frac{epoch}{epoch\_drop}}}
\end{equation}

Here, $lr$ is the current learning rate, $lr_0$ is the initial learning rate, $drop\_rate$ is the factor used for reduction (for this implementation $drop\_rate$ is 0.5), an epoch is the current epoch and $epoch\_drop$ is after how many epochs the learning rate is factored down. Here, $lr$ is the current learning rate, $lr_0$ is the initial learning rate, $drop\_rate$ is the factor used for reduction (for this implementation $drop\_rate$ is 0.5), an epoch is the current epoch and $epoch\_drop$ is after how many epochs the learning rate is factored down.

\begin{figure*}
\includegraphics[width=\textwidth]{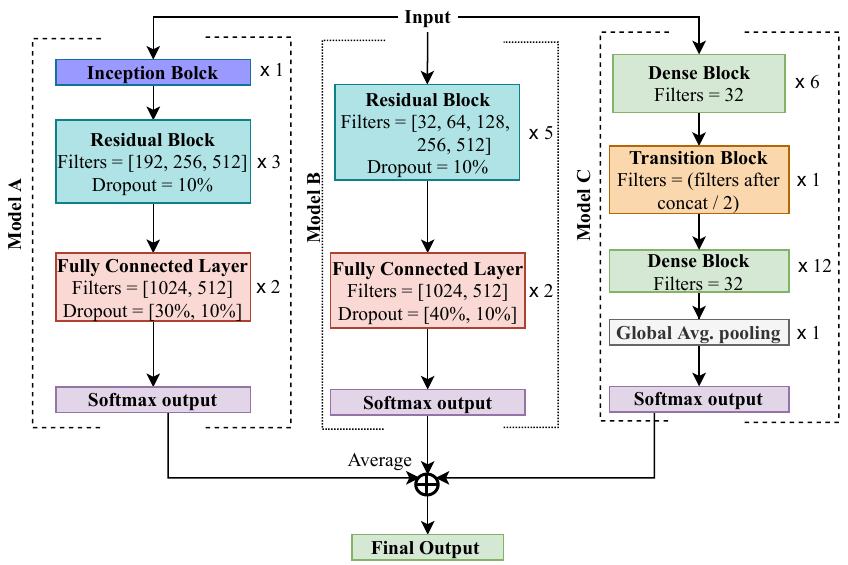}
\caption{Block diagram of BanglaNet (‘Filters’ and ‘Dropout’ is the number of filters and dropout rate used in each consecutive layer)} \label{fig3}
\end{figure*}

\subsection{Inception block}
Different from any other CNN architecture, inception was first proposed by Szegedy et al. \cite{szegedy2015going}. Inception was introduced while keeping in mind the variability of images in a dataset and the effect of different kernel sizes on them. Different shapes of kernel or filter size are appropriate for a variety of images, i.e., centered, more distributed, more locally distributed images, etc. The idea of inception was to use every type of kernel in a single layer, thus making the network deep and wide. Fig. \ref{fig2}(a) depicts an inception block. After performing a convolution operation of kernel sizes (3x3), (5x5), and a pooling operation; all the outputs are concatenated or stacked together. The only problem with this type of operation was concatenating increases the dimension and thus, computation complexity and storage required. To solve this problem (1x1) convolution is used as a bottleneck layer that limits the depth of the layers and reduces the computational cost.  As in Fig. \ref{fig2}(a) an (1x1) convolution is performed before any convolution operation to reduce the shape received from the previous layer. In the original paper, \cite{he2016deep} an additional convolution layer with (1x1) kernel size was concatenated with the other filters and pooling layer, which is omitted for this paper. 

The notion of residual blocks in CNN was portrayed by He et al. in their publication \cite{he2016deep}. Conventionally, CNNs are structured i.e., they are formed by a sequence of blocks containing a convolution layer followed by a pooling layer and ending with fully connected layers. The problem with these types of networks is that with an increasing number of layers, they are more prone to vanishing gradient which makes the performance decrease instead of increasing. To solve this problem and maintain the identity function, ResNet made use of residual blocks where later layers are connected with earlier layers which made it possible to use much deeper layers by avoiding the vanishing gradient problem. Fig. \ref{fig2}(b) delineates a residual block. As can be seen, there is a shortcut connection between the first convolutional layer and the third convolutional layer; it is ensured that the channel length of both of these layers is equal and an element-wise addition operation is performed. This modifies the function after adding non-linearity from $F(x)$ to $F(x)+x$, where, $x$ is identity mapping. By doing so, the residual block makes it easier to make the residuals tend to zero given identity function is optimal, this enables to use deeper model without being distressed by vanishing gradient. Apart from that, ResNet is shown by experimentation to converge much faster than its plain counterpart \cite{he2016deep}. The original ResNet architecture has not made use of dropout but, for the implementation of this paper, dropout is included as the regularization method. Also, in \cite{he2016deep} convolution operation with stride 2 was used to downsample the input which is replaced with average pooling for this paper. 
\begin{table}[htbp]
\caption{Dataset properties}\label{tab1}
\begin{center}
\begin{tabular}{|m{1.5cm}|m{0.7cm}|m{1.3cm}|m{1cm}|m{1cm}|m{0.8cm}|m{1cm}|m{0.8cm}}
\hline
\textbf{Dataset} & \textbf{Total classes} & \textbf{Training image per class} & \textbf{Total training images} & \textbf{Test images per class} & \textbf{Total test images} \\
\hline
\textbf{CMATERdb} & 231 & 220-400 & 49282 & 50-200 & 13254 \\
\hline
\textbf{BanglaLekha-Isolated} & 84 & 1550-1650 & 130378 & 250-300 & 32551 \\
\hline
\textbf{Ekush} & 122 & 2300-2400 & 282835 & 500-600 & 70646\\
\hline
\end{tabular}
\end{center}
\end{table}

\begin{figure*}[h]
\centering
  \subfloat[CMATERdb]{
	\begin{minipage}[c][1\width]{
	   0.3\textwidth}
	   \centering
	   \includegraphics[scale=.35]{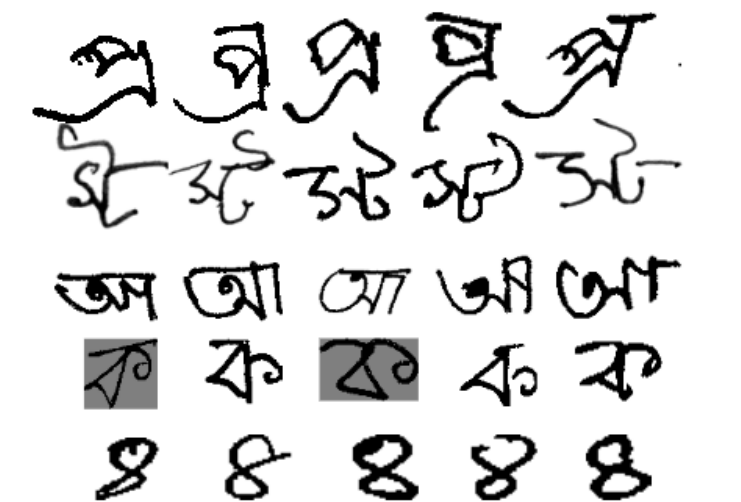}
	\end{minipage}}
 \hfill 	
  \subfloat[BanglaLekha Isolated]{
	\begin{minipage}[c][1\width]{
	   0.3\textwidth}
	   \centering
	   \includegraphics[scale=.3]{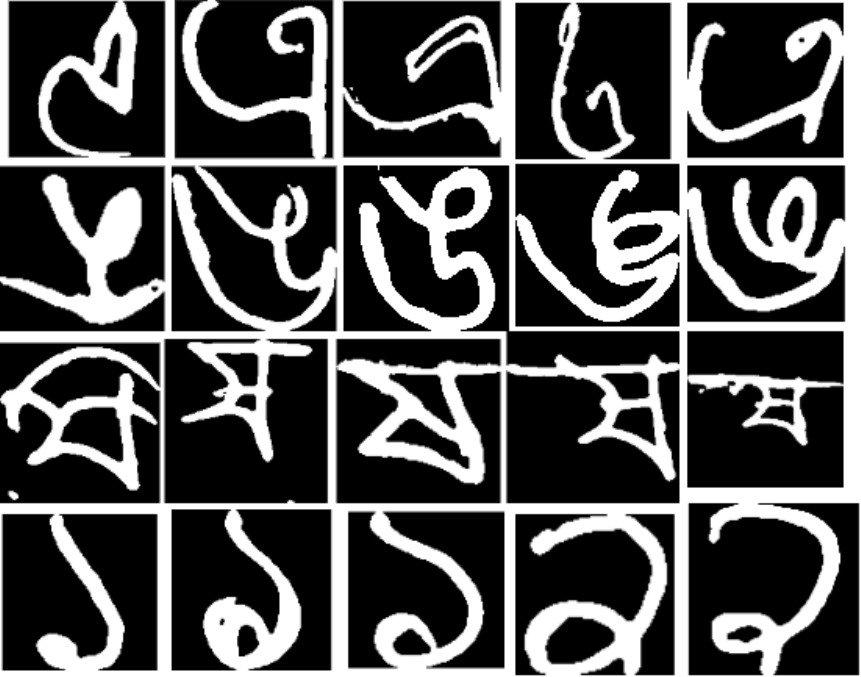}
	\end{minipage}}
 \hfill	
  \subfloat[Ekush]{
	\begin{minipage}[c][1\width]{
	   0.3\textwidth}
	   \centering
	   \includegraphics[scale=.5]{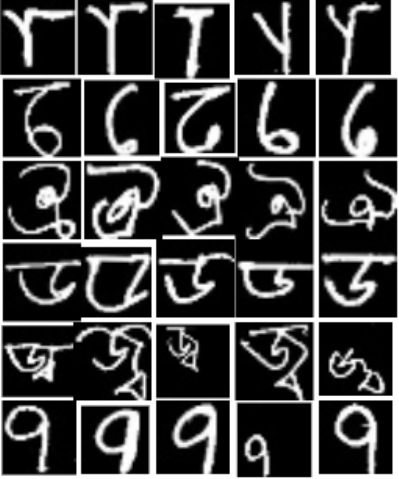}
	\end{minipage}}
\caption{Samples from Datasets}\label{fig4}
\end{figure*}

\subsection{Densely connected convolutional layers}
A densely connected network was proposed by Huang et al. \cite{huang2017densely} which can be thought of as a variation of ResNet. In DenseNet, all convolution layer is connected to all previous convolutional layers. All feature maps from previous layers are concatenated channel-wise instead of adding weights like ResNet. A dense block is displayed in fig \ref{fig2}(c). DenseNet has similar issues as inception; as channel-wise concatenation is performed, the depth of some layers can be large which would increase the computational power needed. To make the model vigorous, DenseNet is divided into two blocks: dense blocks and transition blocks. On the dense block features concatenated together follows a growth rate. Because each layer is getting information from all previous layers, the growth rate or number of feature maps added for each convolutional layer can be shallow. Although the growth rate can be narrowed, a bottleneck layer is used to ensure the network does not exceed a limit of feature maps for computational efficiency. The (1x1) convolutional layer is used to bottleneck the growth. After applying several dense blocks, to ensure the compactness of the model, a transition block containing a (1x1) convolutional layer is used to decrease the number of channels further. For this paper, a growth rate of 32 and compression ratio of 0.5 for the transition layer is maintained, i.e., the channel depth will be halved occasionally.  DenseNet enables deep supervision by allowing the loss function to give direct feedback to the earlier layers during backward propagation. As earlier features are concatenated to later ones, features are diversified allowing a robust and generalized model to be created. 

\begin{table*}[h!]
\caption{experimental results on Test set}\label{tab2}
\begin{center}
\begin{tabular}{|m{1.1cm}|m{1.5cm}|m{1.1cm}|m{1.1cm}|m{1.1cm}|m{1.1cm}|m{1.1cm}|m{1.1cm}|m{1.6cm}|}
\hline
Dataset &Evaluation matrix &\multicolumn{3}{|c|}{Augmented input} &\multicolumn{3}{|c|}{Non-augmented input} &Ensembled: BanglaNet\\
\cline{3-8}
& &Model A &Model B &Model C &Model A &Model B &Model C &\\
\hline
\multirow{3}{*}{\rotatebox{60}{\tiny{CMATERdb}}} &Accuracy &\textbf{98.07}\%	&97.77\%	&97.77\% &97.43\%	&97.46\%	&97.23\% &\textbf{98.40\%}\\ 
\cline{2-9}
  & Top-3   accuracy &\textbf{99.68\%}	&99.64\%	&99.63\% &99.56\%	&99.59\%	&99.51\% &\textbf{99.76\%}\\
\cline{2-9}
  & Loss  &\textbf{0.0731}	&0.0776	&0.0755 &0.1263 &0.1255	&0.1171 & \textbf{0.0562}\\
\hline

\multirow{3}{*}{\rotatebox{60}{\tiny{BanglaLekha}}} &Accuracy &\textbf{97.43\%}	&97.38\%	&97.13\%	&96.88\%	&96.85\%	&96.73\%	&\textbf{97.56\%}\\
\cline{2-9}
  & Top-3   accuracy &\textbf{99.67\%}	&99.66\%	&99.63\%	&99.56\%	&99.52\%	&99.52\%	&\textbf{99.74\%}\\
\cline{2-9}
  & Loss  &0.1092	&\textbf{0.1022}	&0.1118	&0.1806	&0.1725	&0.1864	&\textbf{0.0929}\\
\hline

\multirow{3}{*}{\rotatebox{60}{Ekush}} &Accuracy &\textbf{97.04\%}	&96.90\%	&96.44\%	&96.61\%	&96.51\%	&96.00\%	&\textbf{97.32\%}\\
\cline{2-9}
  & Top-3   accuracy &\textbf{99.47\%}	&99.46\%	&99.36\%	&99.36\%	&99.29\%	&99.05\%	&\textbf{99.56\%}\\
\cline{2-9}
  & Loss  &0.1487	&\textbf{0.1300}	&0.1593	&0.2639	&0.2443	&0.3133	&\textbf{0.1145}\\
\hline
\end{tabular}
\end{center}
\end{table*}

\begin{figure}[h!]
\centering
  \subfloat[Accuracy vs Epochs]{
	\begin{minipage}[c][1\width]{
	   0.4\textwidth}
	   \includegraphics[width=180pt, height=170pt]{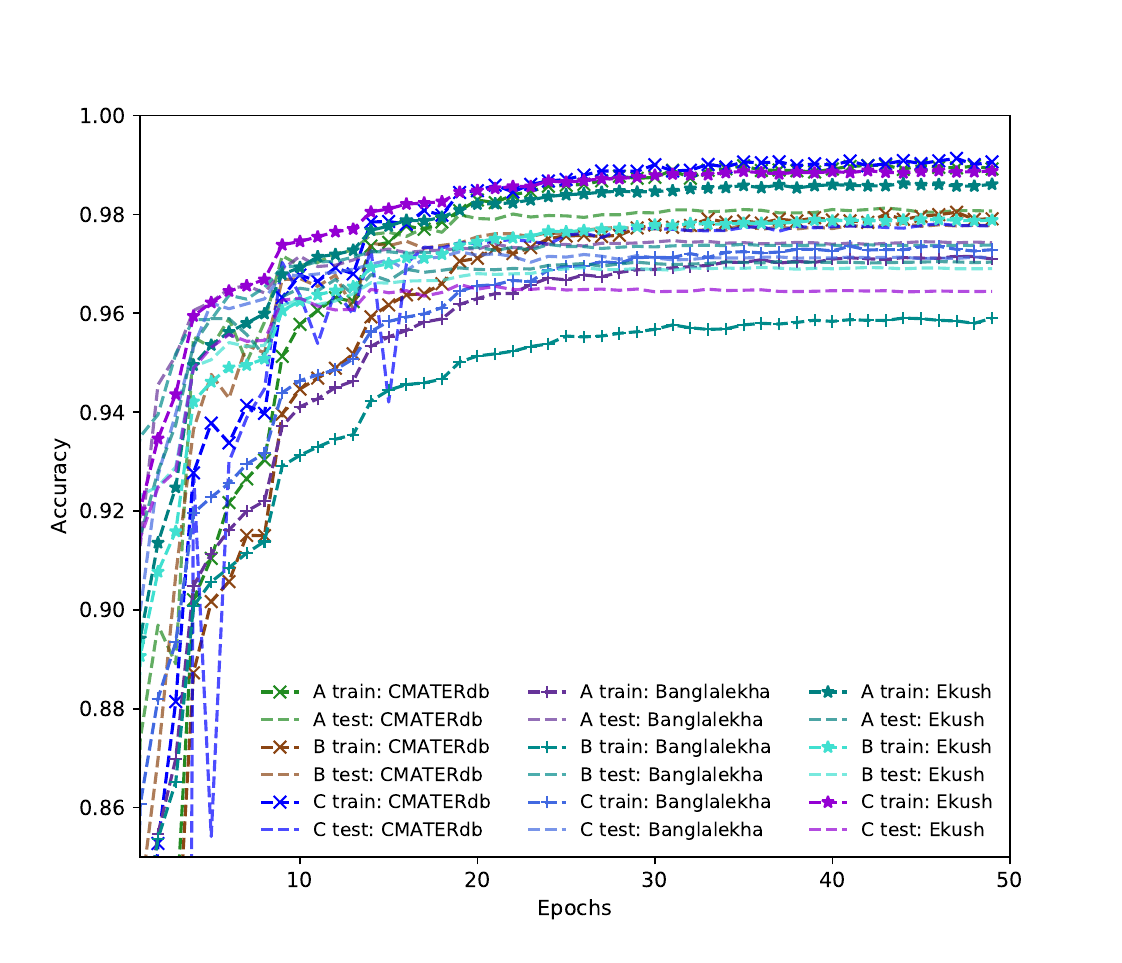}
	\end{minipage}}
\hspace{0.09\textwidth}	
  \subfloat[Loss vs Epochs]{
	\begin{minipage}[c][1\width]{
	   0.4\textwidth}
	   \includegraphics[width=180pt, height=170pt]{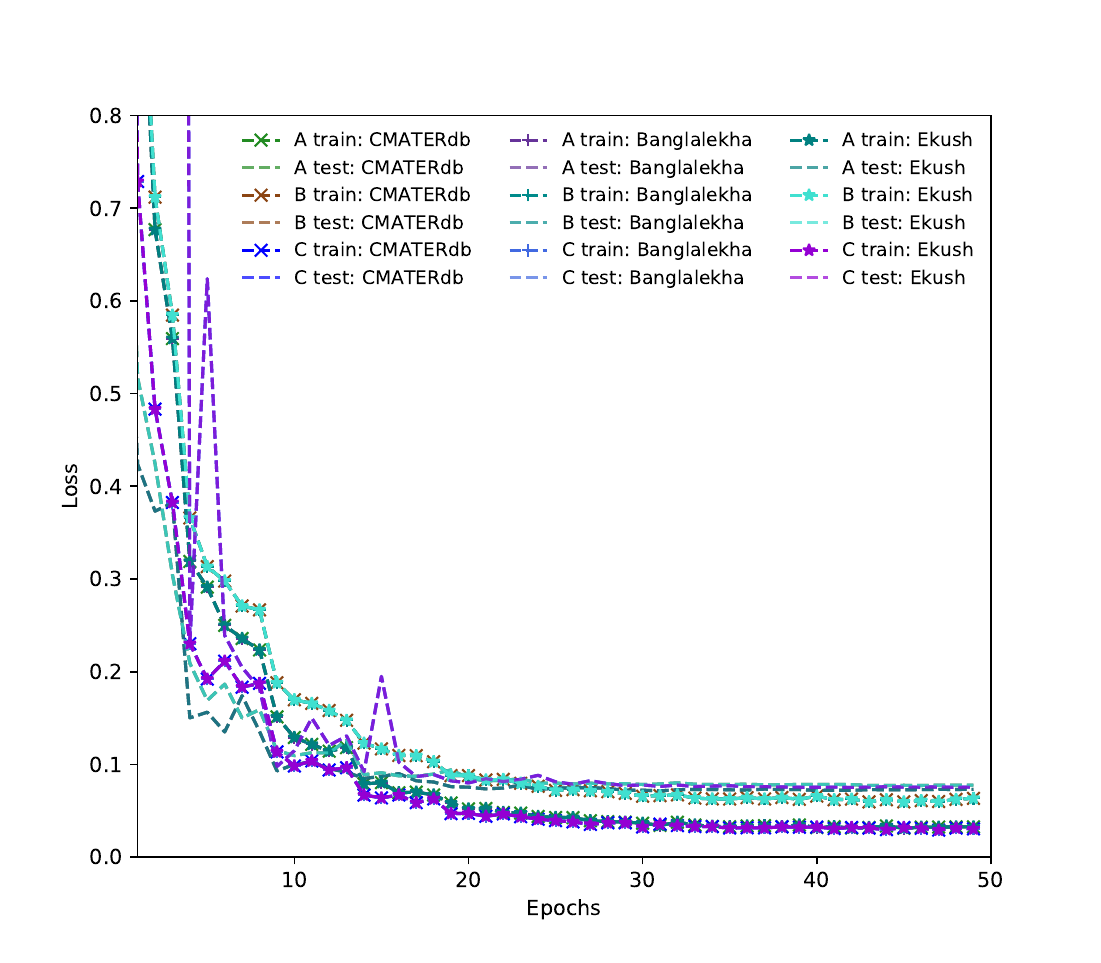}
	\end{minipage}}
\caption{Learning curves for training and validation data}\label{fig5}
\end{figure}

\section{Proposed model: BanglaNet}
The proposed model is constructed by ensembling three different models: A, B, and C. These three models are derived using ideas of resNet \cite{he2016deep}, Inception \cite{szegedy2015going} and DenseNet \cite{huang2017densely}. Original versions of these models require extreme computational resources and a tremendous amount of training time for the huge number of parameters, because of their particularly deep or wide architecture. These models are built for large scale, object detection datasets like ImageNet \cite{deng2009imagenet}. Even with sophisticated resources, it takes weeks, even months to train such models. However finding an optimal CNN model involves tuning parameters, training several models, and selecting the best one. As a result, to build an optimal Bangla isolated handwritten character classifier recognition accuracy as well as computational resource consumption need to be considered. Ensembling different models is a way to boost recognition performance \cite{noor2018handwritten} by decreasing the loss obtained from each separate model in consideration. In ensembling, output predictions obtained from different models are averaged and this average value is taken as the final prediction. Ensembling is especially effective because different models perform well on different sets of test data. Therefore, an improved ensembled model is proposed that gives the best performance for three prominent Bangla isolated character datasets. A detailed block diagram of the model is delineated in Fig. \ref{fig3}.

In Fig. \ref{fig3}, Model A consists of an Inception block (Fig. \ref{fig2}, b) followed by three residual blocks and ending with 2 fully connected layers and a softmax output layer. Several studies have used the idea of inception block followed by simple ‘conv-pool’ and gained good results \cite{rabby2018ekush,rabby2018bornonet,alom2017handwritten}. So, the idea of an inception block with a residual block is explored in this paper. Dropout layers with a drop rate of 20\% are used in each residual block. Model B contains five residual blocks (Fig. \ref{fig2}, a) with 32, 64, 128, 256, and 512 filters for the convolutional layer respectively. Additional dropout layers with a 10\% drop rate are used after each ReLu activation operation. In models A and B, two fully connected layers with 1024 and 512 nodes, with dropout layers are used after flattening the result achieved from the previous residual block. Model C comprises 6 dense blocks (fig. \ref{fig2}, c), accompanied by a transition layer where the number of filters is halved, followed by 12 dense blocks. At the end of the last dense block, a Global Average Pooling (GAP) unit is used where each feature map from the previous layer is averaged producing a single number for each filter, thus a linear array is obtained which is then fed into the softmax output layer. Dropout regularization is not used for model C. 

Categorical cross-entropy is used as a loss function to train all the models and as an optimization technique, Adam is used \cite{kingma2014adam}. The mathematical formula for categorical cross entropy is given in the Equation (\ref{eq3}), it calculates the difference between predicted label $\hat{y}$ and true label $y$ for each training sample with a logarithmic function. 
\begin{equation}\label{eq3}
\scriptsize
H_{\hat{y}} = - \sum_{i} \hat{y}log(y_i)
\end{equation}
As a regularization method, batch normalization, and image augmentation are used in all the models. Additionally, dropout regularization is used in models A and B. As adaptive learning rate step decay is used as discussed earlier. The initial learning rate is ascertained after some trial and error, 0.0005 for models A and B, and 0.001 for model C.

All these models are trained with augmentation and without augmentation separately for the previously discussed datasets. Output from softmax units of each of these models is then ensembled to get the final model, BanglaNet.

\section{Experimental evaluation}
The Proposed model is developed with Keras framework~\cite{chollet2015keras} with TensorFlow as backend~\cite{abadi2016tensorflow} with python as programming language. The hardware configuration used is, an Intel core i5-3470 CPU with 16 GB RAM and 4GB NVIDIA GeForce 1050 Ti GPU. The datasets used and detailed experimental results obtained are discussed in the following subsections.

\subsection{Dataset description}
The experimentation is performed on three prominent and reasonably large datasets designed for Bangla handwritten character recognition. Table~\ref{tab1} delineates some properties of these datasets. All these datasets are inspected manually to correct mislabeled samples and some very confusing samples are removed. This data reconnoiter is done to maintain a strongly labeled dataset. Table~\ref{tab1} contains some properties of the datasets used. detailed description of these datasets with some sample images is given in the following subsections.

\subsubsection{CMATERdb: }
Created at the "Center for Microprocessor Applications for Training Education and Research" research laboratory, CMATERdb is a moderately large dataset collection for Bangla handwritten character recognition. For experimenting three detached versions of this same dataset are integrated. These are CMATERdb 3.1.1 for Bangla numerals, CMATERdb 3.1.1 for Bangla basic characters, and CMATERdb 3.1.3.1 for Bangla compound characters. The total number of classes is 231 including 50 basic, 10 numerals, and 171 compound characters. This is the biggest dataset for its’ purpose, based on the number of classes. Despite being a large dataset, the training and test sample varies severely class-wise. The range of training images per class is 200 to 400 for compound characters, 300 for basic characters, and 400 for digits. Test images per class are roughly in the range of 40 to 70 for compound characters, 60 for basic, and 200 for digits. This diverse number of samples makes it a hard classification problem. The images of the dataset are reasonably clean and centered although, the image size varies. Fig. \ref{fig4}(a) shows some sample images from the dataset.

\subsubsection{BanglaLekha-Isolated: }
Proposed by Biswas et al., BanglaLekha-Isolated is a large dataset for Bangla handwritten characters containing 50 basic characters, 10 digits, and 24 predominantly used compound characters \cite{biswas2017banglalekha} (Fig. \ref{fig4}(b)). This dataset is very diverse as it was collected from people of age 6 to 28 where most of them were 16 to 20 years old. The number of images per class is almost the same containing 1900 to 2000 images. The writing quality of most of the images is specious. Almost every class contained around 10 noisy images; some characters were not recognizable. The entire dataset is examined manually to remove these inferior images. Nevertheless, many sample images are quite hard to identify even for human eyes. Some samples are exhibited in Fig.~\ref{fig4}(b). This data set is perhaps the most difficult one among these three datasets. 80 percent of the entire dataset is selected as training images and 20 percent as test images.

\begin{figure}[ht]
\centering
  \subfloat[CMATERdb]{
	\begin{minipage}[c][1\width]{
	   0.3\textwidth}
	   \centering
	   \includegraphics[scale=.6]{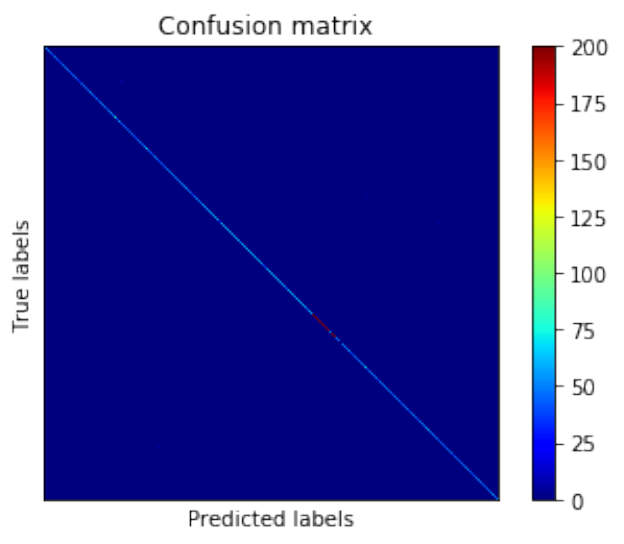}
	\end{minipage}}
 \hfill 	
  \subfloat[BanglaLekha Isolated]{
	\begin{minipage}[c][1\width]{
	   0.3\textwidth}
	   \centering
	   \includegraphics[scale=.6]{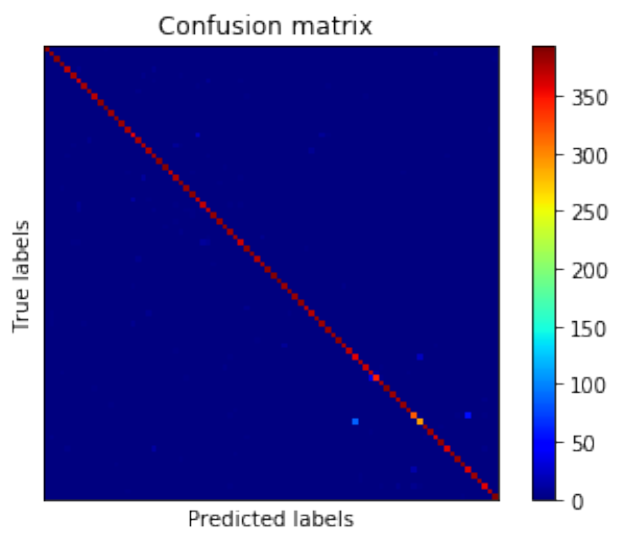}
	\end{minipage}}
 \hfill	
  \subfloat[Ekush]{
	\begin{minipage}[c][1\width]{
	   0.3\textwidth}
	   \centering
	   \includegraphics[scale=.6]{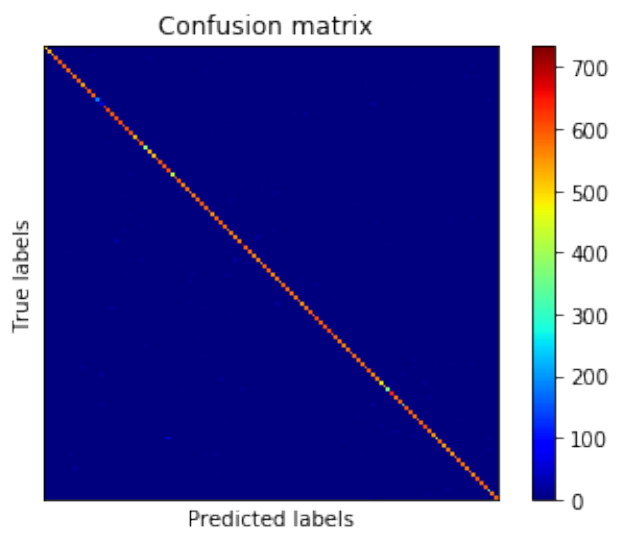}
	\end{minipage}}
\caption{Confusion Matrices for Each Datasets}\label{fig6}
\end{figure}

\begin{table}[htbp]
\caption{Average precision, recall and F1 score obtained for each dataset }\label{tab3}
\begin{center}
\begin{tabular}{|m{1.7cm}|m{1.9cm}|m{1.4cm}|m{1.8cm}|}
\hline
\textbf{Dataset} & \textbf{Avg. precision} & \textbf{Avg. recall} & \textbf{Aug. F1 score}\\
\hline
\textbf{CMATERdb} & 98\% & 98\% & 98\% \\
\hline
\textbf{BanglaLekha-Isolated} & 98\% & 98\% & 98\% \\
\hline
\textbf{Ekush} & 97\% & 97\% & 97\%\\
\hline
\end{tabular}
\end{center}
\end{table}

\begin{table*}[h!]
\caption{Comparative analysis }\label{tab4}
\begin{center}
\begin{tabular}{|m{2.5cm}|m{4cm}|m{3cm}|c|}
\hline
\textbf{Dataset} & \textbf{Work} & \textbf{Number of Classes} & \textbf{Accuracy}\\
\hline
\multirow{5}{*}{\textbf{CMATERdb}} & BornoNet \cite{rabby2018bornonet}	&50 (basic characters)	&98\%\\
\cline{2-4}
& Modified ResNet-18 \cite{alif2017isolated}	&231	&95.99\%\\
\cline{2-4}
& DCNN with ELU and dropout \cite{ashiquzzaman2017efficient}	&171(compound characters)	&93.68\%\\
\cline{2-4}
& DCNN \cite{roy2017handwritten}	&171 (compound characters)	&90.33\%\\
\cline{2-4}
& \textbf{BanglaNet}	&\textbf{231}	&\textbf{98.40\%}\\
\hline
\multirow{6}{*}{\textbf{BanglaLekha}} & BornoNet \cite{rabby2018bornonet}	&50 (basic characters)	&95.71\%\\
& CNN with inception \cite{alom2017handwritten}	&84	&89.3\%\\
\cline{2-4}
& Modified ResNet-18 \cite{alif2017isolated}	&84	&95.10\%\\
\cline{2-4}
& ResNet-50 with transfer learning \cite{chatterjee2019bengali}	&84	&96.12\%\\
\cline{2-4}
&CNN with augmentation \cite{chowdhury2019bangla} &50(Basic characters)	&95.25\%\\
\cline{2-4}
& \textbf{BanglaNet}	&\textbf{84}	&\textbf{97.57\%}\\
\hline
\multirow{2}{*}{\textbf{Ekush}} & EkushNet \cite{rabby2018ekush}	&122	&97.73\%\\
\cline{2-4}
& \textbf{BanglaNet}	&\textbf{122}	&\textbf{97.32\%}\\
\hline
\end{tabular}
\end{center}
\end{table*}
\subsubsection{Ekush:}
Ekush, proposed by Rabby et al. is the biggest dataset based on the size of samples \cite{rabby2018ekush}. The Ekush dataset contains a total of 367,018 images of isolated characters for 122-character classes. Among these classes 50 characters are basic, 10 for digits, 10 modifiers, and 52 frequently used compound characters. The number of images per class is almost the same in this dataset except for two compound character classes. According to the authors of \cite{rabby2018ekush} they primarily think of getting 50 compound characters but while retrieving data from 3086 people, many of them misinterpreted 2 compound characters and wrote other compound characters. These two wrongly written characters were separated and made into two more classes. It can be inferred that; those two compound character classes are the characters with the least number of samples. However, the quality of the writing is quite good and there are moderate variations in writing style. Images are noise-free and the size of all images is very small, 28 x 28 px. Fig. \ref{fig4}(c) shows some example images from the dataset. This is a particularly challenging dataset not only because so many classes of characters are involved but also the existence of some imbalanced classes among compound characters.

\subsection{Result and discussion}
All the models are trained for 50 epochs with augmented and non-augmented data inputs separately with parameter settings discussed in the ‘Proposed model: BanglaNet’ section. All the outputs from these models are finally ensembled. Amongst the datasets, CMATERdb already has training and testing images divided, for the other two datasets 80\%-20\% of images are considered as training and testing images.

Table~\ref{tab2} discussed the results obtained for each dataset. It can be observed the accuracy obtained is much higher and, especially, the loss obtained after ensembling is much lower than even the best-performing model (highlighted in boldface).

Table \ref{tab3} gives more details on the model performance with average precision, recall, and F1 score obtained from each dataset with BanglaNet. In Fig. \ref{fig5} learning curves for each model with augmented data are shown and in Fig. \ref{fig6} confusion matrices indicating the number of correctly or incorrectly predicted samples are given. Upon careful consideration of the results it is apparent that the ensembled CNN model BanglaNet acquired outstanding recognition accuracy for all the datasets which is supported by additional evaluation matrices of Table~\ref{tab3}. To gain additional insights the evaluations are compared with some very recent research works based on CNN. Comparative analysis is shown in Table~\ref{tab4}.

from the comparative analysis of Table \ref{tab4} we can see that the proposed BanglaNet achieved significant performance gain for CMATERdb and BanglaLekha isolated datasets. On the Ekush dataset, the accuracy is marginal with the EkushNet \cite{rabby2018ekush} model. In the case of CMATERdb and BanglaLekha isolated datasets, only the basic characters or compound characters are considered in a lot of instances~\cite{rabby2018bornonet,ashiquzzaman2017efficient,chowdhury2019bangla,roy2017handwritten}.

\section{Conclusion}
\label{sec:conclusion}
An ensemble DCNN model is proposed in this paper that comprises the output average of three different models built on the principles of Inception, ResNet, and DenseNet architecture. The effectiveness of the introduced model, BanglaNet has been shown using conclusive results of remarkable recognition accuracies and reduced complexity on three prominent datasets, namely, CMATERdb, BanglaLekha Isolated, and Ekush. The proposed model outperforms the classification accuracies of numerous very recent models while using all types of Bangla character classes including basic characters, compound characters, modifiers, and numerals. The task of building a complete Bangla Handwritten Character Recognition system with the help of the proposed model, BanglaNet will be addressed in future works.

\balance

\bibliographystyle{IEEEtran}
\bibliography{text/main}

\begin{thebibliography}{10}
\providecommand{\url}[1]{#1}
\csname url@samestyle\endcsname
\providecommand{\newblock}{\relax}
\providecommand{\bibinfo}[2]{#2}
\providecommand{\BIBentrySTDinterwordspacing}{\spaceskip=0pt\relax}
\providecommand{\BIBentryALTinterwordstretchfactor}{4}
\providecommand{\BIBentryALTinterwordspacing}{\spaceskip=\fontdimen2\font plus
\BIBentryALTinterwordstretchfactor\fontdimen3\font minus \fontdimen4\font\relax}
\providecommand{\BIBforeignlanguage}[2]{{%
\expandafter\ifx\csname l@#1\endcsname\relax
\typeout{** WARNING: IEEEtran.bst: No hyphenation pattern has been}%
\typeout{** loaded for the language `#1'. Using the pattern for}%
\typeout{** the default language instead.}%
\else
\language=\csname l@#1\endcsname
\fi
#2}}
\providecommand{\BIBdecl}{\relax}
\BIBdecl

\bibitem{basu2012handwritten}
S.~Basu, N.~Das, R.~Sarkar, M.~Kundu, M.~Nasipuri, and D.~K. Basu, ``Handwritten bangla alphabet recognition using an mlp based classifier,'' \emph{arXiv preprint arXiv:1203.0882}, 2012.

\bibitem{fukushima1982neocognitron}
K.~Fukushima and S.~Miyake, ``Neocognitron: A self-organizing neural network model for a mechanism of visual pattern recognition,'' in \emph{Competition and cooperation in neural nets}.\hskip 1em plus 0.5em minus 0.4em\relax Springer, 1982, pp. 267--285.

\bibitem{lecun1998gradient}
Y.~LeCun, L.~Bottou, Y.~Bengio, and P.~Haffner, ``Gradient-based learning applied to document recognition,'' \emph{Proceedings of the IEEE}, vol.~86, no.~11, pp. 2278--2324, 1998.

\bibitem{krizhevsky2012imagenet}
A.~Krizhevsky, I.~Sutskever, and G.~E. Hinton, ``Imagenet classification with deep convolutional neural networks,'' in \emph{Advances in neural information processing systems}, 2012, pp. 1097--1105.

\bibitem{deng2009imagenet}
J.~Deng, W.~Dong, R.~Socher, L.-J. Li, K.~Li, and L.~Fei-Fei, ``Imagenet: A large-scale hierarchical image database,'' in \emph{2009 IEEE conference on computer vision and pattern recognition}.\hskip 1em plus 0.5em minus 0.4em\relax Ieee, 2009, pp. 248--255.

\bibitem{he2016deep}
K.~He, X.~Zhang, S.~Ren, and J.~Sun, ``Deep residual learning for image recognition,'' in \emph{Proceedings of the IEEE conference on computer vision and pattern recognition}, 2016, pp. 770--778.

\bibitem{szegedy2015going}
C.~Szegedy, W.~Liu, Y.~Jia, P.~Sermanet, S.~Reed, D.~Anguelov, D.~Erhan, V.~Vanhoucke, and A.~Rabinovich, ``Going deeper with convolutions,'' in \emph{Proceedings of the IEEE conference on computer vision and pattern recognition}, 2015, pp. 1--9.

\bibitem{das2012statistical}
N.~Das, J.~M. Reddy, R.~Sarkar, S.~Basu, M.~Kundu, M.~Nasipuri, and D.~K. Basu, ``A statistical--topological feature combination for recognition of handwritten numerals,'' \emph{Applied Soft Computing}, vol.~12, no.~8, pp. 2486--2495, 2012.

\bibitem{das2012novel}
N.~Das, K.~Acharya, R.~Sarkar, S.~Basu, M.~Kundu, and M.~Nasipuri, ``A novel ga-svm based multistage approach for recognition of handwritten bangla compound characters,'' in \emph{Proceedings of the International Conference on Information Systems Design and Intelligent Applications 2012 (INDIA 2012) held in Visakhapatnam, India, January 2012}.\hskip 1em plus 0.5em minus 0.4em\relax Springer, 2012, pp. 145--152.

\bibitem{das2009handwritten}
N.~Das, S.~Basu, R.~Sarkar, M.~Kundu, M.~Nasipuri, and D.~Basu, ``Handwritten bangla compound character recognition: Potential challenges and probable solution.'' in \emph{IICAI}, 2009, pp. 1901--1913.

\bibitem{das2015improved}
N.~Das, S.~Basu, R.~Sarkar, M.~Kundu, M.~Nasipuri \emph{et~al.}, ``An improved feature descriptor for recognition of handwritten bangla alphabet,'' \emph{arXiv preprint arXiv:1501.05497}, 2015.

\bibitem{dasbenchmark}
N.~Das, K.~Acharya, R.~Sarkar, S.~Basu, M.~Kundu, and M.~Nasipuri, ``A benchmark data base of isolated bangla handwritten compound characters,'' \emph{IJDAR (Revised version communicated)}.

\bibitem{biswas2017banglalekha}
M.~Biswas, R.~Islam, G.~K. Shom, M.~Shopon, N.~Mohammed, S.~Momen, and A.~Abedin, ``Banglalekha-isolated: A multi-purpose comprehensive dataset of handwritten bangla isolated characters,'' \emph{Data in brief}, vol.~12, pp. 103--107, 2017.

\bibitem{rabby2018ekush}
A.~S.~A. Rabby, S.~Haque, M.~S. Islam, S.~Abujar, and S.~A. Hossain, ``Ekush: A multipurpose and multitype comprehensive database for online off-line bangla handwritten characters,'' in \emph{International Conference on Recent Trends in Image Processing and Pattern Recognition}.\hskip 1em plus 0.5em minus 0.4em\relax Springer, 2018, pp. 149--158.

\bibitem{rabby2018bornonet}
A.~S.~A. Rabby, S.~Haque, S.~Islam, S.~Abujar, and S.~A. Hossain, ``Bornonet: Bangla handwritten characters recognition using convolutional neural network,'' \emph{Procedia computer science}, vol. 143, pp. 528--535, 2018.

\bibitem{bhattacharya2008handwritten}
U.~Bhattacharya and B.~B. Chaudhuri, ``Handwritten numeral databases of indian scripts and multistage recognition of mixed numerals,'' \emph{IEEE transactions on pattern analysis and machine intelligence}, vol.~31, no.~3, pp. 444--457, 2008.

\bibitem{alom2017handwritten}
M.~Z. Alom, P.~Sidike, T.~M. Taha, and V.~K. Asari, ``Handwritten bangla digit recognition using deep learning,'' \emph{arXiv preprint arXiv:1705.02680}, 2017.

\bibitem{alom2018handwritten}
M.~Z. Alom, P.~Sidike, M.~Hasan, T.~M. Taha, and V.~K. Asari, ``Handwritten bangla character recognition using the state-of-the-art deep convolutional neural networks,'' \emph{Computational intelligence and neuroscience}, vol. 2018, 2018.

\bibitem{abir2019bangla}
B.~Abir, S.~N. Mahal, M.~S. Islam, and A.~Chakrabarty, ``Bangla handwritten character recognition with multilayer convolutional neural network,'' in \emph{Advances in Data and Information Sciences}.\hskip 1em plus 0.5em minus 0.4em\relax Springer, 2019, pp. 155--165.

\bibitem{alif2017isolated}
M.~A.~R. Alif, S.~Ahmed, and M.~A. Hasan, ``Isolated bangla handwritten character recognition with convolutional neural network,'' in \emph{2017 20th International Conference of Computer and Information Technology (ICCIT)}.\hskip 1em plus 0.5em minus 0.4em\relax IEEE, 2017, pp. 1--6.

\bibitem{ashiquzzaman2017efficient}
A.~Ashiquzzaman, A.~K. Tushar, S.~Dutta, and F.~Mohsin, ``An efficient method for improving classification accuracy of handwritten bangla compound characters using dcnn with dropout and elu,'' in \emph{2017 Third International Conference on Research in Computational Intelligence and Communication Networks (ICRCICN)}.\hskip 1em plus 0.5em minus 0.4em\relax IEEE, 2017, pp. 147--152.

\bibitem{hakim2019handwritten}
S.~A. Hakim \emph{et~al.}, ``Handwritten bangla numeral and basic character recognition using deep convolutional neural network,'' in \emph{2019 International Conference on Electrical, Computer and Communication Engineering (ECCE)}.\hskip 1em plus 0.5em minus 0.4em\relax IEEE, 2019, pp. 1--6.

\bibitem{chatterjee2019bengali}
S.~Chatterjee, R.~K. Dutta, D.~Ganguly, K.~Chatterjee, and S.~Roy, ``Bengali handwritten character classification using transfer learning on deep convolutional neural network,'' \emph{arXiv preprint arXiv:1902.11133}, 2019.

\bibitem{kingma2014adam}
D.~P. Kingma and J.~Ba, ``Adam: A method for stochastic optimization,'' \emph{arXiv preprint arXiv:1412.6980}, 2014.

\bibitem{chowdhury2019bangla}
R.~R. Chowdhury, M.~S. Hossain, R.~ul~Islam, K.~Andersson, and S.~Hossain, ``Bangla handwritten character recognition using convolutional neural network with data augmentation,'' in \emph{2019 Joint 8th International Conference on Informatics, Electronics \& Vision (ICIEV) and 2019 3rd International Conference on Imaging, Vision \& Pattern Recognition (icIVPR)}.\hskip 1em plus 0.5em minus 0.4em\relax IEEE, 2019, pp. 318--323.

\bibitem{rizvi2019comparative}
M.~A.~I. Rizvi, K.~Deb, M.~I. Khan, M.~M.~S. Kowsar, and T.~Khanam, ``A comparative study on handwritten bangla character recognition,'' \emph{Turkish Journal of Electrical Engineering \& Computer Sciences}, vol.~27, no.~4, pp. 3195--3207, 2019.

\bibitem{noor2018handwritten}
R.~Noor, K.~M. Islam, and M.~J. Rahimi, ``Handwritten bangla numeral recognition using ensembling of convolutional neural network,'' in \emph{2018 21st International Conference of Computer and Information Technology (ICCIT)}.\hskip 1em plus 0.5em minus 0.4em\relax IEEE, 2018, pp. 1--6.

\bibitem{roy2017handwritten}
S.~Roy, N.~Das, M.~Kundu, and M.~Nasipuri, ``Handwritten isolated bangla compound character recognition: A new benchmark using a novel deep learning approach,'' \emph{Pattern Recognition Letters}, vol.~90, pp. 15--21, 2017.

\bibitem{hasan2019bangla}
M.~J. Hasan, M.~F. Wahid, and M.~S. Alom, ``Bangla compound character recognition by combining deep convolutional neural network with bidirectional long short-term memory,'' in \emph{2019 4th International Conference on Electrical Information and Communication Technology (EICT)}.\hskip 1em plus 0.5em minus 0.4em\relax IEEE, 2019, pp. 1--4.

\bibitem{ioffe2015batch}
S.~Ioffe and C.~Szegedy, ``Batch normalization: Accelerating deep network training by reducing internal covariate shift,'' \emph{arXiv preprint arXiv:1502.03167}, 2015.

\bibitem{JMLR:v15:srivastava14a}
\BIBentryALTinterwordspacing
N.~Srivastava, G.~Hinton, A.~Krizhevsky, I.~Sutskever, and R.~Salakhutdinov, ``Dropout: A simple way to prevent neural networks from overfitting,'' \emph{Journal of Machine Learning Research}, vol.~15, no.~56, pp. 1929--1958, 2014. [Online]. Available: \url{http://jmlr.org/papers/v15/srivastava14a.html}
\BIBentrySTDinterwordspacing

\bibitem{huang2017densely}
G.~Huang, Z.~Liu, L.~Van Der~Maaten, and K.~Q. Weinberger, ``Densely connected convolutional networks,'' in \emph{Proceedings of the IEEE conference on computer vision and pattern recognition}, 2017, pp. 4700--4708.

\bibitem{chollet2015keras}
F.~Chollet \emph{et~al.}, ``Keras: Deep learning library for theano and tensorflow,'' \emph{URL: https://keras. io/k}, vol.~7, no.~8, p.~T1, 2015.

\bibitem{abadi2016tensorflow}
M.~Abadi, P.~Barham, J.~Chen, Z.~Chen, A.~Davis, J.~Dean, M.~Devin, S.~Ghemawat, G.~Irving, M.~Isard \emph{et~al.}, ``Tensorflow: A system for large-scale machine learning,'' in \emph{12th $\{$USENIX$\}$ symposium on operating systems design and implementation ($\{$OSDI$\}$ 16)}, 2016, pp. 265--283.

\end{thebibliography}

\vspace{12pt}

\end{document}